\pdfoutput=1

\documentclass[11pt]{article}

\usepackage[]{ACL2023}

\usepackage{times}
\usepackage{latexsym}
\usepackage{booktabs}
\usepackage{longtable}
\usepackage{graphicx}
\usepackage{xcolor}
\usepackage{textcomp}
\usepackage{tabularx}
\usepackage{array}

\usepackage{times}
\usepackage{latexsym}
\usepackage{tabularx}
\usepackage{multirow}
\usepackage{xspace}
\usepackage{booktabs}
\usepackage{tabularray}
\UseTblrLibrary{booktabs}
\usepackage{makecell}
\usepackage{lipsum}
\usepackage{enumitem}
\usepackage{graphicx}
\usepackage{calc}
\usepackage{fdsymbol}
\usepackage[T1]{fontenc}

\usepackage[utf8]{inputenc}

\usepackage{microtype}

\usepackage{inconsolata}

\title{\methodnosp:
Understanding Self-Contradictions in Documents with Large Language Models}

\author{Jierui Li\textsuperscript{$\clubsuit$}$^*$\hspace{15pt} Vipul Raheja\textsuperscript{$\diamondsuit$}  \hspace{15pt} Dhruv Kumar\textsuperscript{$\diamondsuit$} \\ \textsuperscript{$\clubsuit$}The University of Texas at Austin \hspace{10pt} \textsuperscript{$\diamondsuit$}Grammarly\\
\texttt{jierui@cs.utexas.edu}, \texttt{\{vipul.raheja,dhruv.kumar\}@grammarly.com}}

\newcommand{\methodnosp}{\textsc{ContraDoc}}
\newcommand{\method}{\textsc{ContraDoc} \hspace{1pt}}
\newcommand{\dataposnosp}{\textsc{ContraDoc-Pos}}
\newcommand{\datapos}{\textsc{ContraDoc-Pos} \hspace{1pt}}
\newcommand{\datanegnosp}{\textsc{ContraDoc-Neg}}
\newcommand{\dataneg}{\textsc{ContraDoc-Neg} \hspace{1pt}}

\begin{document}
\maketitle
\begin{abstract}
In recent times, large language models (LLMs) have shown impressive performance on various document-level tasks such as document classification, summarization, and question-answering. However, research on understanding their capabilities on the task of self-contradictions in long documents has been very limited. In this work, we introduce \methodnosp, the first human-annotated dataset to study self-contradictions in long documents across multiple domains, varying document lengths, self-contradiction types, and appearance scope. We then analyze the current capabilities of four state-of-the-art open-source and commercially available LLMs: GPT3.5, GPT4, PaLM2, and LLaMAv2 on this dataset.  While GPT4 performs the best and can outperform humans on this task, we find that it is still unreliable and struggles with self-contradictions that require more nuance and context. We release the dataset \footnote{\url{https://github.com/ddhruvkr/CONTRADOC}} 
and all the code associated with the experiments.
\let\thefootnote\relax\footnote{$^*$Work done while Jierui was an intern at Grammarly.}
\end{abstract}

\section{Introduction}


Detecting contradictions in texts has long been pivotal in natural language understanding(NLU), with most of the works falling under the umbrella of natural language inference(NLI)\cite{10.5555/1597538.1597659, 10.1007/11736790_9, de-marneffe-etal-2008-finding}. Detecting contradictions is often regarded as determining the relation between a hypothesis and a piece of premise. However, understanding contradictions when they occur within the confines of a single text (self-contradictions), and furthermore, doing so holistically at the document-level, is still under-explored\cite{9671319}. 
\begin{figure}[!t]
  \centering
  \includegraphics[width=\linewidth]{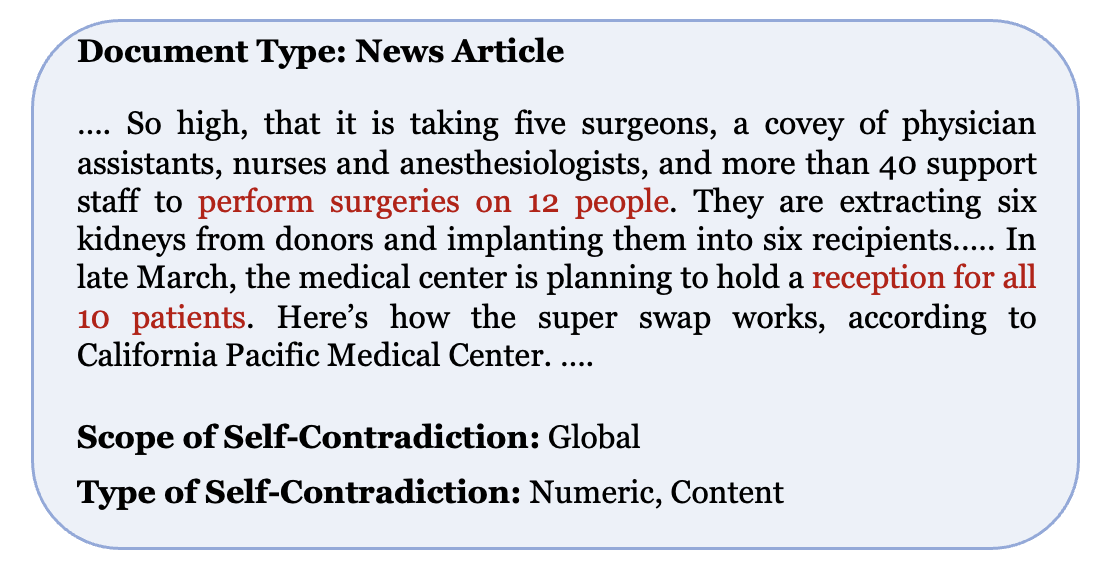}
  \caption{Example of a self-contradictory document from \methodnosp. The highlighted parts in green show the evidence for the self-contradiction. Additionally, information about the scope and type of the contradiction is also present.}
  \label{fig:dataset_ex}
\end{figure}

A text is considered self-contradictory when it contains multiple ideas or statements that inherently conflict. This could manifest in various ways, such as the existence of logical paradoxes, antithetical assertions, or inconsistent descriptions. Figure \ref{fig:dataset_ex} shows an example of self-contradiction in a document. The highlighted two sentences provide contradictory information about the number of patients, thus resulting in a self-contradictory document. 

Psychological research \cite{graesser1993anomalous, doi:10.1111/j.1467-9280.1992.tb00034.x} indicates that humans struggle to identify contradictions in unfamiliar, informative texts, particularly when contradictions are widely separated in long documents, underscoring the need for automated text analysis tools to tackle this challenge.


Previous research on document-level contradictions either focused on sentence-document pair NLI\cite{yin2021docnli, schuster2022stretching} or has been restricted to a single type of document\cite{9671319}. \citet{9671319} defined self-contradiction detection as a binary classification task, proving inadequate for accurately evaluating since they do not require locating self-contradictions within texts. 

To further explore the study in this domain, we propose a new document-level self-contradictory dataset \method with the following characteristics:
\begin{itemize}[noitemsep]
    \item The documents are from different sources and of different lengths. 
    \item The documents and the highlighted self-contradictions within are automatically generated and verified by human annotators. 
    \item It contains a variety of self-contradictions, with each contradiction tagged with information such as its type and appearance scope by human annotators. 
    \item The resulting self-contradictory documents are contextually fluent, thus, keeping the document coherent and plausible. 
\end{itemize}



To create \methodnosp, we utilize a human-machine collaborative framework. We first use LLMs and NLP pipelines to automatically create and introduce self-contradiction into a consistent document. Then, human annotators verify and label attributes for the self-contradictory documents, ensuring the quality and utility of our dataset.

The advent of large language models (LLMs) pre-trained on extensive context lengths \cite{NEURIPS2020_1457c0d6, chowdhery2022palm} has shown promising results over various document-level tasks spanning document classification\cite{sun2023text}, document summarization\cite{zhang2023benchmarking}, document-level question answering\cite{singhal2023expertlevel}, and document-level machine translation\cite{wang2023documentlevel}. 
Yet, we argue that LLMs’ abilities to handle tasks with long context are inconsistent, given their significant dependence on the specific characteristics of the task. 
To investigate how well can large language models detect self-contradiction in documents, we evaluate state-of-the-art, open-source and commercially available LLMs: GPT3.5\cite{OpenAI_ChatGPT_Blog}, GPT4\cite{gpt4openai}, PaLM2\cite{anil2023palm}, and LLaMAv2\cite{touvron2023llama} on \methodnosp. 

We design three evaluation tasks and corresponding metrics to assess LLMs' zero-shot performance. In our experiments, we find that even SOTA models cannot achieve applicable performance. We did a thorough study on the effects of different aspects of documents and self-contradictions and found that LLMs can detect object self-contradictions(e.g., facts) much better than subject self-contradictions(e.g., emotion or perspective). 


In summary, this paper makes the following contributions:
\begin{itemize}[noitemsep]
    \item We propose a human-annotated dataset consisting of self-contradictory documents across varying document domains and lengths and self-contradiction types and appearance scope, being the first work to touch on those aspects.  
    \item We propose three evaluation tasks and corresponding metrics to evaluate the performance of models on detecting self-contradictions in text. The proposed evaluation goes beyond binary judgment and focuses on the models' ability to pinpoint self-contradictions. 
    \item We conduct an extensive analysis of four of the best-performing LLMs (open-source or commercially available) and provide insights into their capabilities of long-text reasoning, focusing on self-contradiction detection in documents. 
\end{itemize}
\begin{table*}[ht]
    \centering
    \small
    \begin{tabular}{p{0.15\linewidth}|p{0.25\linewidth}|p{0.25\linewidth}|p{0.25\linewidth}} 
     \toprule
     \textbf{Type} & \textbf{Definition} & \textbf{Original Statement} & \textbf{Generated Self-Contradiction} \\ 
     \midrule
     Negation & Negating the original sentence  & Zully donated her kidney. & Zully never donated her kidney. \\ 
     \midrule
     Numeric & Number mismatch or number out of scope. & All the donors are between 20 to 45 years old. & Lisa, who donates her kidney, she is 70 years old. \\ 
     \midrule
     Content & Changing one/multiple attributes of an event or entity & Zully Broussard donated her kidney to a stranger. & Zully Broussard donated her kidney to her close friend. \\
     \midrule
     Perspective / View / Opinion & Inconsistency in one’s attitude/ perspective/opinion & The doctor spoke highly of the project and called it “a breakthrough” & The doctor disliked the project, saying it had no impact at all. \\
     \midrule
     Emotion / Mood / Feeling & Inconsistency in one’s attitude/ emotion/mood & The rescue team searched for the boy worriedly. & The rescue team searched for the boy happily. \\
     \midrule
     Relation & Description of two mutually exclusive relations between entities. & Jane and Tom are a married couple. & Jane is Tom’s sister. \\
     \midrule
     Factual & Need external world knowledge to confirm the contradiction. & The road T51 was located in New York. & The road T51 was located in California. \\
     \midrule
     Causal & The effect does not match the cause. & I slam the door. & After I do that, the door opens. \\
     \bottomrule
    \end{tabular}
    \caption{Definition and example of sentence rewriting for different types of self-contradictions.}
    \label{tab:type_of_contra_ex}
\end{table*}

\section{Related Work}

\subsection{Detecting Contradictions in Text}
The problem of detecting contradictory statements in texts has been long explored in NLP literature \cite{condoravdi-etal-2003-entailment, 10.5555/1597538.1597659}, mainly as a text classification or textual entailment task. 
Most prior work has studied contradictions under the Natural Language Inference (NLI) framework of evaluating contradictory pairs of sentences, namely, as Recognizing Textual Entailment (RTE) tasks \cite{10.1007/11736790_9, bowman-etal-2015-large}. Contradiction detection has also been explored in dialogue \citet{nie-etal-2021-like, zheng-etal-2022-cdconv, jin-etal-2022-improving}, question answering systems \cite{fortier-dubois-rosati-2023-using}.  

More recently, a fair amount of NLI research has focused on long-document reasoning, going beyond sentence-level granularity to document-level,\cite{yin-etal-2021-docnli, schuster-etal-2022-stretching, mathur-etal-2022-docinfer}. However, these works differ from ours as they either frame the tasks as NLI, do not focus on investigating the capabilities of LLMs, or do not focus on self-contradictions. 

Contradiction detection has been investigated in various other domains, such as Social Media \cite{lendvai-reichel-2016-contradiction, lendvai-etal-2016-monolingual, 8508308} for detecting rumorous posts on Twitter or in user opinions in Amazon product reviews; or to detect and fix contradictions in Financial \cite{deusser2023contradiction} or Biomedical \cite{ROSEMBLAT2019103275, sarafraz2012finding, Alamri2016, alamri2016detection} reports.


\subsection{Understanding Self-Contradictions}
Despite the extensive amount of research into studying contradictions, there has been a very limited amount of work that has focused on self-contradictions in long documents. The closest work to ours is \citet{9671319} on Wikipedia-based contradiction detection, where they curated a dataset based on the "Self-contradictory" template on Wikipedia and used a pairwise model to detect it. \methodnosp dataset significantly differs from their proposed dataset in the variety of document types, contradiction types and additional annotations it contains.
\citet{mündler2023selfcontradictory} refine LLM-generated texts to eliminate contradictions, both relevant yet distinct from our comprehensive, domain-inclusive approach focusing on holistic document analysis with LLMs. 


\section{\method Dataset}

\method contains 449 self-contradictory (referred to as \dataposnosp) and 442 non-contradictory documents (referred to as \datanegnosp). Non-contradictory documents are defined as documents that do not contain any self-contradictions and are considered negative examples for the task. We include them in our dataset to evaluate if the models can identify the documents that do not contain any self-contradictions sampled from the same source of contradictory documents. 
Furthermore, the documents in \method cover three domains, vary in length and scope of dependencies, and contain different types of contradictions. This allows us to see how these variations affect the performance of the LLMs. In the development of our dataset, we leverage a human-machine collaborative framework, where human experts evaluate and verify machine-generated self-contradictions, ensuring the created data is both rich and reliable. We only use documents written in English in this work.

\subsection{Dataset Statistics}
\label{sec:dataset-statistics}

The overall statistics for the 449 documents in \dataposnosp are shown later in this paper in Table \ref{tab:fine_grained}. The distribution of non-contradictory documents in \dataneg is similar to \datapos.

The different attributes of our dataset pertaining to self-contradiction types, document, and context lengths, and the research questions used to study them are outlined below.

\paragraph{RQ1: Are self-contradictions harder to detect in some domains for LLMs?}
To create \methodnosp, we construct a document corpus from three domains to test the performance in various contexts. We use CNN-DailyMail dataset \cite{hermann2015teaching} for news articles, NarrativeQA \cite{kovcisky2018narrativeqa} for stories, and WikiText \cite{merity2016pointer} for Wikipedia documents (details in Appendix \ref{sec:dataset_details}). For each document, one self-contradiction is inserted in.

\paragraph{RQ2: Are self-contradiction harder to detect in longer documents for LLMs?} Documents in \methodnosp range from 100 tokens to 2200 tokens helping us study both longer and shorter documents. 
Table \ref{tab:fine_grained} shows the detailed breakdown of our dataset with respect to document lengths (in tokens). 

\paragraph{RQ3: Are self-contradictions present farther away in a document more difficult to detect for LLMs?}
To test the effect of context length on the model's performance, we introduce contradictions that are present at different distances from each other. 
We define the appearance scope as follows: The instances where contradictions are present within a sentence are labeled as \textit{intra}, whereas the instances where the contradictory statements are present four sentences or less apart are labeled \textit{local}, and finally, the instances where the contradictions are present more than four sentences apart are labeled \textit{global}. Our dataset contains 73, 220, and 155 documents with intra, local, and global contradictions. \label{sec:scope}

\paragraph{RQ4: Are some types of self-contradictions harder to detect than others for LLMs?}  

\citet{de-marneffe-etal-2008-finding} defined contradictions into two broad categories, content and lexical, \citet{10.1145/3477495.3531881} defined six types of self-contradictions similarly for sentence-level contradiction detection. 

We focus on the content and extend it to build a more fine-grained taxonomy. introduce a more complete choice of types to study: 
Each document in \methodnosp is tagged with one or multiple of the following eight types of self-contradictions: Negation, Numeric, Content, Perspective/View/Opinion, Emotion/Mood/Feeling, Factual, Relation, and Causal. 

A more comprehensive overview is presented in \ref{tab:type_of_contra_ex} with examples. The exact table is also provided for our annotators to annotate the dataset.


\begin{figure}[!t]
  \centering
  \includegraphics[width=\linewidth]{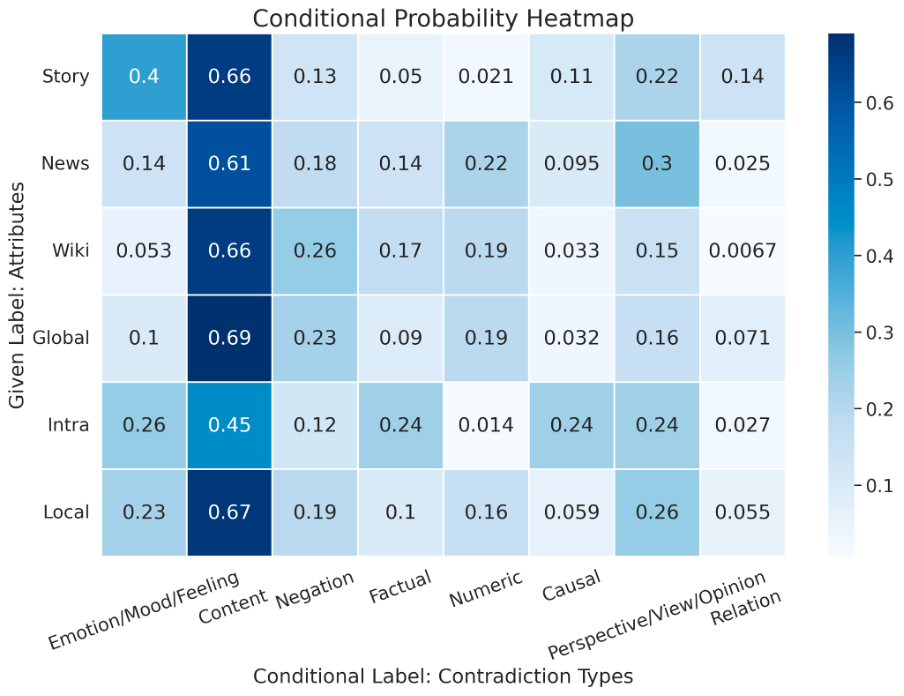}
  \caption{Label dependencies, shown with conditional probabilities. Each cell is the occurrence probability of the x-axis label, given the presence of the y-axis label.}
  \label{fig:heatmap}
  \vspace*{-2mm}
\end{figure}

The labeled attributes in our dataset are not independent of each other. We illustrate the conditional probabilities over the contradiction types and other properties in Figure \ref{fig:heatmap} to show the pairwise dependencies. For the self-contradiction type, ``Content'' is the most common type as it often co-occurs with other types like ``Negation'', ``Numeric'' or ``Factual''. 40\% of story documents contain ``Emotion/Mood/Feeling'' self-contradictions while this number is only ``14\%'' and ``5.3\%'' for news and wiki. This indicates that the distributions of types of self-contradictions vary amongst different types of documents. This should be considered and we analyze the more fine-grained performance on different labels in experiments section \ref{sec:fine-grained}.










    

\subsection{Dataset Creation Method}

\begin{figure}[!t]
  \centering
  \includegraphics[width=\linewidth]{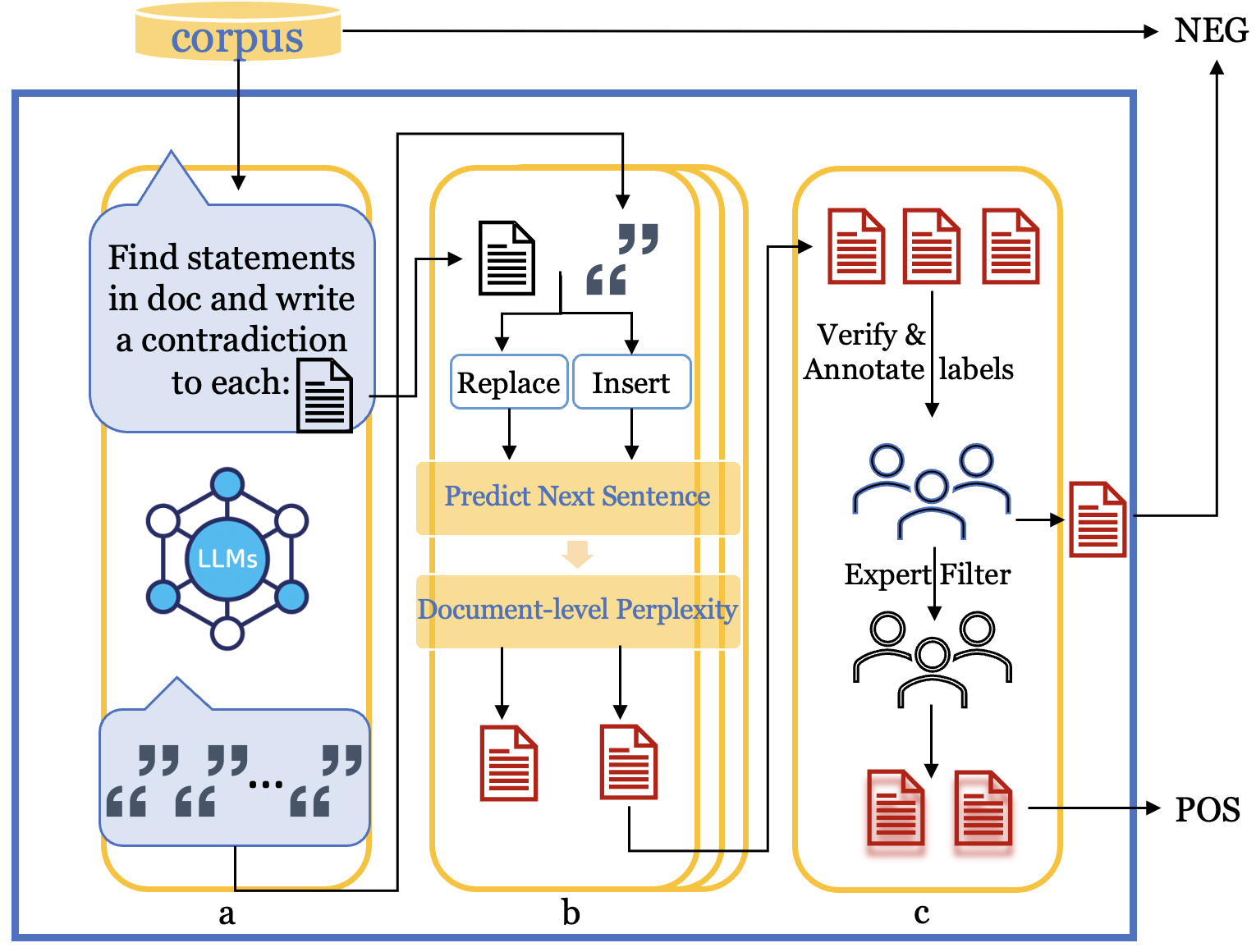}
  \caption{Dataset Creation Pipeline. a) Contradictory Statements Generation using LLMs; b) Self-Contradictory Document Creation; c) Human verification and Tagging.}
  \label{fig:method}
\vspace*{-2mm}
\end{figure}

While LLMs are used widely in data labeling and dataset creation \cite{ding-etal-2023-gpt1, Wang2021WantTR}, \citet{Pangakis2023AutomatedAW} argues that the data annotated by generative AI requires human verification. Thus, we utilize a human-machine collaborative framework to create our dataset. We first automatically create and introduce self-contradictions into a document. Then, we ask human annotators to verify and label attributes for the contradictory documents. The data creation process is systematically organized into three primary components: a) Contradictory Statements Generation; b) Self-Contradictory Document Creation; c) Human Verification and Tagging. Figure \ref{fig:method} provides an overview of the dataset creation process. 


\subsubsection{Contradictory Statements Generation Using LLM }
Given an initially consistent document $d$ that doesn't contain self-contradiction, we process it through an LLM (GPT-4-0314 in our case) to generate contradictory statements by asking it to identify $k$ statements $st_{1}, st_{2}, \cdots, st_{k}$ in the document and generate a contradictory statement to each of the $k$ statements, yielding $k$ contradictions correspondingly: $c_{1}, c_{2}, \cdots, c_{k}$. More specifically, we provide few-shot examples of contradictory statements of different types, guiding the LLM to identify and generate more diverse statements.

In practice, the model tends to edit only a few words in the statement unless explicitly asked otherwise. To make contradictory statements sound natural, we also ask it to rephrase it using a different wording $c'_{1}, c'_{2}, \cdots, c'_{k}$. Thus for a single document provided, LLM generates $k$ triplets: $(st_{i}, c_{i}, c'_{i})$

\subsubsection{Self-Contradictory Document Creation}
Upon obtaining $k$ of  $(st_{i}, c_{i}, c'_{i})$ triplets, we modify the source document by either \textit{inserting} the contradictory statement $c_{i}\text{ or } c'_{i}$ in the document or \textit{replacing} the original statement $st_{i}$ with $c_{i}\text{ or } c'_{i}$, forming a candidate set of potentially contradictory documents $\hat{D_i} = \{\hat{d_i}(ins-c_i), \hat{d_i}(ins-c'_i), \hat{d_i}(rep-c_i), \hat{d_i}(rep-c'_i)\}$. This is driven by two assumptions: 1) Introducing contradictory facts separately may render the document self-contradictory. 2) Directly substituting statements with contradictory versions might induce contextual inconsistency. 

To maintain document fluency while introducing contradiction, we apply the following metrics to filter in self-contradictory documents from the candidate set: 
\begin{itemize}[itemsep=-2pt]
    \item \textbf{Global Fluency}: We measure document-level perplexity and ensure that it does not exceed a defined threshold, $T$, post-editing.
\begin{equation}
\begin{split}
\text{\small $ppl(d) = \exp\left(\frac{1}{n}\right) \sum_{j=1}^n \log(P(w_j))$}\\
\text{\small $ppl(\hat{d}_i) - ppl(d) \leq T$}
\end{split}
\end{equation}
where $n$ is the total number of tokens in document $d$ and $P(w_j)$ is the probability to predict token $w_j$. In practice, we set $T=0.01 \text{ to } 0.03$ for different types and lengths of documents. 
    
    \item \textbf{Local Fluency}: We employ BERT’s ``Next Sentence Prediction(NSP)'' task \cite{devlin-etal-2019-bert} to validate the contextual coherence of the modified sentences. After placing the modified sentence in $c_i$ or $c'_i$ at position $j$ th, we accept such edit if: NSP$(s_{j-1}, s_j)$ and NSP$(s_j, s_{j+1})$ are both True. 
\end{itemize}
If multiple contradictory documents in $ \hat{D_i}$ meet the mentioned constraints, we accept the one with the lowest global perplexity to maintain diversity in self-contradictions.

\subsubsection{Human Verification and Tagging}

An additional human annotation layer was integrated to validate the automated modifications, ensuring the resultant documents were both natural and genuinely contradictory. We highlight the original statement as well as the introduced self-contradiction in the document as Figure \ref{fig:dataset_ex} for annotators\footnote{The annotators were native English speakers from the US with at least a Bachelor's degree in English.} to verify the validity of document-level self-contradiction as well as tagging labels for self-contradiction type and scope of self-contradiction(intra, local, global as in Section \ref{sec:scope}). The questions can be found in \ref{sec:anno}. 

Each modified document was evaluated by two annotators, establishing consensus on the self-contradiction and document validity. 
Examples are filtered if both annotators verify that the modification makes a valid document-level self-contradiction. When annotators disagree, we select ``closer'' option for self-contradiction scope while joining different self-contradiction types. 

To verify the annotation quality, we run another expert filter by the authors of this work to verify controversial cases marked by annotators. Regarding the self-contradiction injection method, the final \method contains 271 documents created by contradictory statement replacing and 178 documents created by contradictory statement inserting. 


\subsubsection{Negative Examples}
We consider the documents without self-contradictions as negative examples in our experiments. While the documents from our source domain can naturally serve as negative examples, we also add modified documents that both annotators tag as ``non-contradictory,'' indicating such modification does not introduce document-level self-contradiction.



\section{Evaluation}

\subsection{Evaluation Tasks and Metrics}

We now describe the evaluation tasks and metrics for different experiments. We design three evaluation tasks, ranging from the simple ``answer Yes or No'' to the more complex ``first judge, then give evidence''. Our experiments and evaluation prompts are designed on respective evaluation tasks. The corresponding prompts for all three experimental settings are in Appendix \ref{sec:prompts}.

\subsubsection{Binary Judgment} \label{sec: bj}
\paragraph{Task} The most straightforward way to evaluate the models is to test their abilities to distinguish between positive and negative examples. We do this by simply asking the model to provide a judgment on whether a document $d$ is self-contradictory or not. In this setting, we evaluate the model on \method.

\paragraph{Prompt Design} We formalize this as the \textit{Binary Judgment} task: Given a document, we ask the model if the document contains a self-contradiction. The model must answer with either "Yes" or "No".

\paragraph{Evaluation Metrics} As \method has balanced positive and negative cases, we use the standard Precision, Recall, F1 score, and Accuracy metrics to evaluate the models' binary judgment, notated as $j(d)$. 

\subsubsection{Self-Contradiction Top-$k$}
\label{sec: topk}
\paragraph{Task} In the zero-shot setting, model performance on the specified task is influenced by its sensitivity to self-contradictions. An under-sensitive model may overlook non-essential self-contradictions, whereas an over-sensitive model could misinterpret minor inconsistencies as contradictions. To address this, we introduce a task aimed at detecting self-contradictions by giving the top $k$ evidential texts. While the self-contradiction introduced by our creation process is assumed to be the most obvious error in the document, it should appear within the top $k$ evidence texts the model provides. We tested on \datapos only. 

\paragraph{Prompt Design}
We formalize this as the \textit{Self-Contradiction Top-$k$}: Given a document with a self-contradiction, we ask the model to select the five most probable sentences that indicate the self-contradiction and rank them from high to low probability. We state in the prompt that the given document contains one self-contradiction. 

\paragraph{Evaluation Metric} We regard \textit{``picking the modified sentence''} as \textit{``finding the self-contradiction''}. A self-contradiction in the document is introduced by either inserting or replacing $c_i$ or $c'_i$, and all other texts are originally in the consistent document $d$. Thus, removing the modified sentence ($c_i$ or $c'_i$) would eliminate the self-contradiction in $\hat{d_i}$. We define $c_i$ or $c'_i$ as the oracle evidence $e_i$. Ideally, the model should also pick another sentence that contradicts $e_i$, but it isn't necessarily the same evidence by annotators as our introduced modification might conflict with many different places in the document.

The evidence of self-contradiction selected by the model must contain the corresponding $e_i$. Instead of doing an exact substring match, we use BertScore \cite{sun-etal-2022-bertscore} to accommodate minor mis-copying: if one of the evidence sentences selected by the model matches $e_i$ with a BertScore Precision > 0.98 or Recall > 0.98, we consider them the same sentence. To verify the evidences $E = \{s_j \mid j = 1, \ldots, k\}$ found by the model, the verification function $v(E)$ is given by: 
\begin{equation}
\scalebox{0.8}{
$\begin{aligned}
v(E) &= 
\begin{cases} 
\text{True} & \text{if } \exists s \in E \text{ such that } \\
& \max(\textsc{BertScore}(s, e_i)_{\text{Prec.}}, \\
& \quad \textsc{BertScore}(s, e_i)_{\text{Rec.}}) > 0.98 \\
\text{False} & \text{otherwise}
\end{cases}
\end{aligned}$
}
\label{eq:equation}
\end{equation}

\noindent We define \textit{Evidence Hit Rate} (EHR) as the percentage of cases where the model could find the correct evidence. In practice, we choose $k=5$ for top $k$. We calculate the EHR to represent the fraction of $v(E)=$ True for \dataposnosp.

\subsubsection{Judge then Find}\label{sec: jtf}
\paragraph{Task} Another drawback with Binary Judgment is that answering ``Yes'' does not necessarily mean the model can find the self-contradiction. We design another task requiring giving not only binary judgment but also the evidence sentence for self-contradiction when answering ``Yes''. In this setting, the model is evaluated on \method.

\paragraph{Prompt Design} We formalize the \textit{Judge-then-Find} task as follows: Given a document, the model needs to determine whether the document has self-contradictions by answering ``Yes" or "No." If the answer is Yes, the model also needs to provide supporting evidence by quoting sentences that can indicate the self-contradiction in the document.

\paragraph{Evaluation Metric} In addition to the metrics mentioned in Section \ref{sec: bj}, the \textit{Verification} $v(E)$ in equation \ref{eq:equation}.
where $k=2$ for E here. 
The \textit{Evidence Hit Rate} (EHR) here is defined as the percentage of cases where the model could find the correct evidence when it answered "Yes". We measure EHR by automatically verifying the supporting evidence provided by the LLMs. It is evaluated only on TPs in this setting, and we show the real accuracy $R-acc (pos)$ over the positive subset \datapos to represent the fraction of $j(d) \land v(E) =$ True. 

\begin{table}[]
    \centering
    \small
    \begin{tabular}{l|c|c|c|c} 
     \toprule
     \textbf{Model} & \textbf{Accuracy} & \textbf{Precision} & \textbf{Recall} & \textbf{F1} \\ 
     \midrule
     GPT3.5 & 50.1\% & 100.0\% & 0.2\% & 0.4 \%\\ 
     GPT4 & 53.8\% & 97.0\% & 8.0\% & 15.6\% \\ 
     PaLM2 & 52.0\% & 61.0\% & 13.4\% & 22.0\% \\
     LLaMAv2 & 50.5\% & 51.0\%& 38.3\% & 43.7\% \\
     \bottomrule
    \end{tabular}
    \caption{Performance of different LLMs on \textbf{Binary Judgement} experiment.} 
    \label{tab:prompt1}
\end{table}

\begin{table}[]
    \centering
    \small
    \begin{tabular}{l|c|c} 
     \toprule
     \textbf{Model} & \textbf{EHR $\uparrow$} & \textbf{Avg. Index (1-5)} $\downarrow$ \\ 
     \midrule
     GPT3.5 & 42.8\% & 1.98 \\ 
     GPT4 & \textbf{70.2}\% & \textbf{1.79} \\ 
     PaLM2 & 48.2\% & 2.36 \\
     LLaMAv2 & 20.4\% & 2.28 \\
     \bottomrule
    \end{tabular}
    \caption{Performance comparison of different LLMs on \textbf{Self-Contradiction in top-$k$} experiment. 
    Evidence Hit Rate(EHR) by random is 16\%. Avg. Index (1-5) is the average index among the top-5 evidence texts where the self-contradiction was found.}
    \label{tab:prompt2}
\end{table}

\begin{table*}
\centering
\small
  \begin{tabular}{>{\centering}p{0.1\linewidth}|>{\centering}p{0.08\linewidth}|>{\centering}p{0.05\linewidth}|>{\centering}p{0.05\linewidth}|>{\centering}p{0.05\linewidth}|>{\centering}p{0.05\linewidth}|>{\centering}p{0.05\linewidth}|p{0.05\linewidth}|>{\centering}p{0.1\linewidth}|p{0.1\linewidth}}
    \toprule
    \textbf{Models} & \textbf{Precision} & \textbf{Recall} & \textbf{F1 Score} & \textbf{TP} rate & \textbf{FP} rate & \textbf{TN} rate & \textbf{FN} rate & \textbf{Evidence Hit Rate} & \textbf{R-acc(pos)}\\
    \midrule
    GPT3.5 & 57.0\% & 62.0\% & 41.0\% & 20.6\% & 12.8\% & 36.9\% & 29.7\% & 41.0\% & 16.8\% \\
    GPT4 & 88.0\% & 39.0\% & 54.0\% & 19.6\% & 2.7\% & 46.2\% & 31.5\% & 92.7\% & 35.6\% \\
    PaLM2 & 52.0\% & 83.0\% & 64.0\% & 41.5\% & 37.6\% & 12.0\% & 9.0\% & 41.0\% & 33.7\% \\
    LLaMAv2 & 50.0\% & 95.0\% & 65.0\% & 48.0\% & 48.6\% & 1.12\% & 2.3\% & 14.5\% & 13.8\% \\
    \bottomrule
  \end{tabular}
  \caption{Performance comparison of different LLMs on \textit{Judge then Find} experimental setting. 
  \textbf{Precision, Recall, F1} and \textbf{TP, FP, TN, and FN} rates are calculated on the entire dataset before verification, i.e., on "Yes/No" prediction. 
  \textbf{Evidence Hit Rate} is the percentage of cases where the model could find the correct evidence when it answered "Yes". \textbf{R-acc(pos)} denotes the fraction of positive data points confirmed by 'yes' judgments and evidence hits.}
\label{tab:prompt3}
\end{table*}


\begin{table*}
\centering
\small
  \begin{tabular}{>{\centering}p{0.1\linewidth}|>{\centering}p{0.08\linewidth}|>{\centering}p{0.08\linewidth}|>{\centering}p{0.08\linewidth}|>{\centering}p{0.08\linewidth}|>{\centering}p{0.2\linewidth}|p{0.12\linewidth}}
    \toprule
    \textbf{Models} & \textbf{TP} rate & \textbf{FP} rate & \textbf{TN} rate & \textbf{FN} rate& \textbf{Evidence Hit Rate} & \textbf{R-acc(pos)} \\
    \midrule
    Human & 18.0\% & 6.7\% & 43.3\% & 32.0\% & 74.1\% & 26.7\% \\
    \midrule
    GPT3.5 & 20.7\% & 15.3\% & 34.7\% & 29.3\% & 25.8\% & 10.7\% \\
    GPT4 & 20.0\% & 4.7\% & 45.3\% & 30.7\% & 86.7\% & 34.7\% \\
    \bottomrule
  \end{tabular}
  \caption{Performance comparison of humans and different LLMs on \textit{Judge then Find} experimental setting on a subset containing 75 positive documents and 75 negative documents. The metrics are similar to those in Table \ref{tab:prompt3}.} 
\label{tab:human_prompt3}
\end{table*}

\subsection{Automatic Evaluation results}

Table \ref{tab:prompt1} shows the results for the \textit{Binary Judgment} Task. We find that all models struggle with detecting self-contradictory documents and predict "No" for most documents, as shown by the low recall values. We observe that LLaMAv2 achieves higher numbers only because it tends to predict ``Yes'' while other models tend to predict ``No'' for most of the cases. The accuracy on the entire dataset, i.e., \datapos and \datanegnosp, is around 50\%, suggesting that the models have a near-random performance. 

Table \ref{tab:prompt2} shows the results for the \textit{Self-Contradiction Top-$k$} Task, where, given a self-contradictory document, the models need to refer to the top-5 probable sentences that can imply the self-contradiction. We find that GPT4 outperforms the other models by a big margin and can correctly detect 70\%  of self-contradictions. PaLM2 is better than GPT3.5 and can correctly detect self-contradictions in 48\% of the documents compared to 43\%. Finally, LLaMAv2 performs the worst and can detect self-contradictions in only 20\% of the documents. We also find that, on average, GPT4 can find the evidence at the 1.79th position out of 5, showing that it is not only best at finding the evidence sentences but also prioritizing them. Note that for all models, the average index that the evidence is found $< 3$, which indicates that the models do rank the evidence by probability of self-contradiction. 
We also provide a deeper analysis in Section \ref{sec:fine-grained}.

Finally, Table \ref{tab:prompt3} shows the results for the \textit{Judge then Find} experiment. In the first part of the task, i.e., answering if the document is self-contradictory or not, similar to results in Table \ref{tab:prompt1}, we find that PaLM2 and LLaMAv2 have a greater bias to answer ``Yes", compared to the GPT models. This is seen in the high TP and FP rates of the two models. However, the low Evidence Success Rates indicate that the models fail to locate the correct evidence when they answer "Yes" to a self-contradictory document. LLaMAv2, in particular, can only find the correct evidence 14.5\% of the time, while GPT3.5 and PaLM2 find correct evidence 41\% of the time. Even though GPT4 might only be able to find 19.6\% of the \dataposnosp, it can provide the correct evidence for 92.7\% of them. GPT4 performs the best in terms of real accuracy, followed closely by the PaLM2 model. 
In summary, we present the following key observations:
\begin{itemize}[itemsep=-1pt]
    \item GPT4 performs the best overall, whereas LLaMAv2 performs the worst.
    \item PaLM2 and LLaMAv2 are biased to answer Yes more often on yes/no prompts, whereas GPTs provide a more balanced output. However, all four models struggle with the yes/no prompts.
    \item While GPT4 predicts ``yes'' less than other models, the evidence hit rate of GPT4 is significantly higher than others, which shows that it is conservative and only answers ``yes'' when being certain about the self-contradiction.
\end{itemize}



\subsection{Human Performance}

We construct a balanced set of documents from our dataset with 150 documents in total and evaluate humans'
performance on the \textit{Judge then Find} task.
Each document is evaluated by one annotator\footnote{The annotators for this task are different from those who worked to verify documents before}. We then also compare their performance with the performance of GPT3.5 and GPT4 on the same documents. Table \ref{tab:human_prompt3} shows the performance comparison. We use the same metrics as the \textit{Judge then Find} experimental setting. 

We find that overall, humans perform better than GPT3.5 but not GPT4. Specifically, we find that humans are the worst at finding TP cases. However, they are much better than GPT3.5 at finding the self-contradiction evidence and does not point out false self-contradiction. 

A possible reason for humans' poor performance is that humans might fail to keep track of details when the document is long, making them miss some self-contradictions. This is a different setting from the annotator verification process, where two potentially contradictory sentences are highlighted, which makes the task easier for humans.

\subsection{Ablation Study}

\label{sec:fine-grained}

\begin{table}[!t]
\centering
\small
  \begin{tabular}{l|c|c|c|c}
    \toprule
    \textbf{Categories} & \textbf{Attributes} &
    \textbf{\# docs} & \textbf{GPT3.5} &
    \textbf{GPT4}\\
    \midrule
    Overall & - & 449  & 42.8\% & 70.2\% \\
    \midrule
    \multirow{3}{*}{\textbf{\shortstack[l]{Document\\Type}}} & \makecell[tc]{news\\wiki\\story} & \makecell[tc]{158\\150\\141}  & \makecell[tc]{45.6\%\\48.0\%\\34.0\%} & \makecell[tc]{65.8\%\\82.0\%\\62.4\%}\\
    \midrule
    \multirow{4}{*}{\shortstack[l]{\textbf{Document}\\\textbf{Length}}} & \makecell[tc]{100-500\\500-1000\\1000-1500\\1500-2200} & \makecell[tc]{50\\184\\143\\72}& \makecell[tc]{50.0\%\\40.2\%\\44.1\%\\41.7\%} & \makecell[tc]{64.0\%\\69.6\%\\74.1\%\\68.1\%} \\
    \midrule
    \multirow{3}{*}{\shortstack[l]{\textbf{Self-}\textbf{Contra}\\\textbf{Scope}}} & \makecell[tc]{global\\local\\intra} & \makecell[tc]{155\\220\\73}  & \makecell[tc]{51.0\%\\38.6\%\\37.0\%} & \makecell[tc]{89.0\%\\63.2\%\\50.7\%} \\
    \midrule
    \multirow{8}{*}{\shortstack[l]{\textbf{Self-}\textbf{Contra}\\\textbf{Type}}} & \makecell[tc]{Negation \\Numeric\\Content\\P/V/O\\E/M/F\\Factual\\Relation\\Causal} & \makecell[tc]{87\\65\\288\\101\\86\\54\\25\\36}  & \makecell[tc]{56.3\%\\58.5\%\\43.4\%\\25.7\%\\29.1\%\textbf{*}\\40.7\%\\40.0\%\\33.3\%} & \makecell[tc]{85.1\%\\87.7\%\\74.7\%\\61.4\%\\50.0\%\\66.7\%\\72.0\%\\55.6\%} \\
    \bottomrule
  \end{tabular}
  \caption{Fine-grained performance of different LLMs on top-$k$ judgment. The scores denote the Evidence Hit Rate. Numbers marked with an asterisk (*) denote Evidence Hit Rate is not statistically significant against random with p-value > 0.05. P/V/O refers to Perspective/View/Opinion while E/M/F refers to Emotion/Mood/Feeling.}
\label{tab:fine_grained}
\end{table}

We now discuss the fine-grained analysis of various models' outputs to get a deeper understanding of their performance on the task of self-contradiction detection and answer the research questions mentioned in Section \ref{sec:dataset-statistics}. We choose the model outputs of GPT3.5 and GPT4 from the \textbf{Self-Contradiction Top-$k$} experimental setting for this analysis. 
We use the probability (p-value) of finding equivalent successes in a binomial test to show the statistical significance of the results against random selecting $k$ sentences from the document. Table \ref{tab:fine_grained} shows the EHR of these models in detecting the self-contradictory statement given in the document. 

\paragraph{RQ1} Among the three document types, we find that models have the highest EHR on Wikipedia documents, followed by News and Stories. GPT4 can detect the self-contradictory statements in 82\% of the Wikipedia documents, compared to 48\% of the cases for GPT3.5. For Stories, the evidence hit rate of GPT4 and GPT3.5 drops to 62.4\% and 34.04\%, respectively. 

\paragraph{RQ2} For both GPT3.5 and GPT4, there is no significant drop in EHR as the document length increases or the other way around. This suggests that the document length is not the main factor determining model's ability to detect self-contradictions. However, documents with relatively short lengths (100-500 tokens) are easier for GPT3.5 to detect the self-contradiction within. 

\paragraph{RQ3} We find that for both GPT3.5 and GPT4, ``global'' self-contradictory documents had a higher EHR than ``local'' and ``intra''. This is in contradiction to our hypothesis that self-contradiction with evidence texts far away might be harder. 
This can be due to label dependencies shown in Figure \ref{fig:heatmap} (discussed ahead). 

\paragraph{RQ4} As we consider the types of self-contradiction types, we find that more objective self-contradiction types, like Numeric and Negation, are the easiest to detect, while more subjective ones like Emotion/Mood/Feeling and Perspective/View/Opinion are hard. We argue this might be because LLMs are pre-trained on more fact-checking tasks aiming to verify facts compared to emotion-consistency tasks.

\paragraph{Dataset Label Dependencies} The fine-grained results in Table \ref{tab:fine_grained} can also be attributed to the label dependencies shown in Figure \ref{fig:heatmap}. As mentioned before, Wikipedia documents are more likely to contain Negation, Numeric and Factual self-contradictions, whereas Stories are more likely to contain Emotion/Mood/Feeling and Perspective/View/Opinion self-contradictions. Similarly, the performance differences in different scopes(global/local/intra) might also be attributed to their distributions of contradiction types. Here, we argue that the models' performance is more related to the self-contradiction type instead of where the self-contradiction is presented or the type of the document.

\section{Additional Sensitivity Analysis}
\label{sec:sensitivity}
\paragraph{Effect of Prompts}
Since we enforce model outputs to a fixed format, this might negatively affect the model performance. This is more true for GPT3.5 than GPT4, which has better instruction-following capability. 
Thus, for 75 documents with self-contradictions, we ask GPT3.5 to generate predictions without putting constraints on the output format (prompt in Appendix \ref{sec:prompts}) and ask humans to evaluate the responses. For 26.4\% cases, it answers ``No''; for 45.8\% of the cases, it provides incorrect evidence; only for 27.8\% of the cases is it able to find the correct evidence (alongside other incorrect evidence). This suggests that the model performance is still far from satisfactory. 
Figure \ref{fig:eval_case} in the Appendix shows the GPT-3.5 outputs for the two cases.

\paragraph{Detecting self-contradictory sentence}
Since we observe that models find it hard to find contradictions in a document, we evaluate the model's capability on an easier task to find a statement that directly contradicts a given sentence. 
Since our dataset contains documents that contain a pair of contradictory sentences, we provide the evidence sentence to the model and ask it to find the contradictory sentence in the document. GPT3.5 can detect 51.6\% of the cases, while GPT4 can detect 77.2\% of them. Such results suggest that LLMs do reasonably well in document-level contradiction detection if the exact sentence with contradiction is pointed out but not so otherwise, but perform much worse in finding self-contradiction if the exact sentence isn't pointed out for its reference. 

\section{Conclusion}

In this work, we present one of the first steps in investigating the task of document-level self-contradictions. We create \methodnosp, a well-annotated dataset for this task, which contains 449 self-contradictory documents spanning over three domains and containing multiple types of self-contradictions. The dataset is annotated by humans and contains information about the scope and type of self-contradiction as well as the evidence to detect self-contradictions. We then investigate the capabilities of four 
state-of-the-art LLMs, namely, GPT3.5, GPT4, PaLM2, and LLaMAv2, on this dataset. We find that overall, GPT4 performs the best and even outperforms humans on the task. However, we also find that there is still a long way to go before GPT4 can reliably detect self-contradictions. We release this dataset and all the associated code for the community to use and develop better document-level reasoning capabilities in LLMs. As part of future work, we plan to investigate the capabilities of LLMs to fix the self-contradictions in the documents.

\section*{Acknowledgement}
We sincerely thank Philip Dwelle and all annotators for their help with the data annotation process. We also thank our colleagues at Grammarly for their helpful comments. We thank all the reviewers, meta-reviewers, and area chairs for their time, efforts, and valuable suggestions on this work. This work was supported by Grammarly.

\section*{Limitations}
Our aim was to create a dataset of self-contradictory documents that sound natural. However, as all self-contradictions are created and inserted automatically, the self-contradictory documents do not always mimic how humans make mistakes or introduce self-contradictions, even though we use humans-in-the-loop. Another limitation is that for some self-contradiction types, we only collected limited data points; for example, there are only 25 documents with \textit{Relation} self-contradictory type in our dataset. Finally, in this work, we only study self-contradictions in English, and our dataset contains documents that are written in English. 

\section*{Ethics Impact}
We propose ContraDoc to encourage attention to the task of self-contradiction, a crucial area that has been notably overlooked in previous research. This task holds substantial practical value in real-world applications like document understanding, evaluation and quality. Moreover, this task has potential applications in legal and academic document analysis, where identifying contradictions can be critical. It's important to clarify that our goal is to augment the capabilities of human professionals, not to replace them. We propose an annotated dataset with automatic evaluation metrics can be a valuable asset to the NLP community, enabling the development and testing of new AI algorithms in this space. Since we build upon fully open-source datasets, we do not see it having any potential risks or negative ethical issues.

\bibliography{anthology,custom}

\begin{thebibliography}{46}
\expandafter\ifx\csname natexlab\endcsname\relax\def\natexlab#1{#1}\fi

\bibitem[{Alamri(2016)}]{alamri2016detection}
Abdulaziz Alamri. 2016.
\newblock \emph{The detection of contradictory claims in biomedical abstracts}.
\newblock Ph.D. thesis, University of Sheffield.

\bibitem[{Alamri and Stevenson(2016)}]{Alamri2016}
Abdulaziz Alamri and Mark Stevenson. 2016.
\newblock \href {https://doi.org/10.1186/s13326-016-0083-z} {A corpus of
  potentially contradictory research claims from cardiovascular research
  abstracts}.
\newblock \emph{Journal of Biomedical Semantics}, 7(1):36.

\bibitem[{Anil et~al.(2023)Anil, Dai, Firat, Johnson, Lepikhin, Passos,
  Shakeri, Taropa, Bailey, Chen et~al.}]{anil2023palm}
Rohan Anil, Andrew~M Dai, Orhan Firat, Melvin Johnson, Dmitry Lepikhin,
  Alexandre Passos, Siamak Shakeri, Emanuel Taropa, Paige Bailey, Zhifeng Chen,
  et~al. 2023.
\newblock Palm 2 technical report.
\newblock \emph{arXiv preprint arXiv:2305.10403}.

\bibitem[{Bowman et~al.(2015)Bowman, Angeli, Potts, and
  Manning}]{bowman-etal-2015-large}
Samuel~R. Bowman, Gabor Angeli, Christopher Potts, and Christopher~D. Manning.
  2015.
\newblock \href {https://doi.org/10.18653/v1/D15-1075} {A large annotated
  corpus for learning natural language inference}.
\newblock In \emph{Proceedings of the 2015 Conference on Empirical Methods in
  Natural Language Processing}, pages 632--642, Lisbon, Portugal. Association
  for Computational Linguistics.

\bibitem[{Brown et~al.(2020{\natexlab{a}})Brown, Mann, Ryder, Subbiah, Kaplan,
  Dhariwal, Neelakantan, Shyam, Sastry, Askell, Agarwal, Herbert-Voss, Krueger,
  Henighan, Child, Ramesh, Ziegler, Wu, Winter, Hesse, Chen, Sigler, Litwin,
  Gray, Chess, Clark, Berner, McCandlish, Radford, Sutskever, and
  Amodei}]{NEURIPS2020_1457c0d6}
Tom Brown, Benjamin Mann, Nick Ryder, Melanie Subbiah, Jared~D Kaplan, Prafulla
  Dhariwal, Arvind Neelakantan, Pranav Shyam, Girish Sastry, Amanda Askell,
  Sandhini Agarwal, Ariel Herbert-Voss, Gretchen Krueger, Tom Henighan, Rewon
  Child, Aditya Ramesh, Daniel Ziegler, Jeffrey Wu, Clemens Winter, Chris
  Hesse, Mark Chen, Eric Sigler, Mateusz Litwin, Scott Gray, Benjamin Chess,
  Jack Clark, Christopher Berner, Sam McCandlish, Alec Radford, Ilya Sutskever,
  and Dario Amodei. 2020{\natexlab{a}}.
\newblock \href
  {https://proceedings.neurips.cc/paper_files/paper/2020/file/1457c0d6bfcb4967418bfb8ac142f64a-Paper.pdf}
  {Language models are few-shot learners}.
\newblock In \emph{Advances in Neural Information Processing Systems},
  volume~33, pages 1877--1901. Curran Associates, Inc.

\bibitem[{Brown et~al.(2020{\natexlab{b}})Brown, Mann, Ryder, Subbiah, Kaplan,
  Dhariwal, Neelakantan, Shyam, Sastry, Askell et~al.}]{brown2020language}
Tom Brown, Benjamin Mann, Nick Ryder, Melanie Subbiah, Jared~D Kaplan, Prafulla
  Dhariwal, Arvind Neelakantan, Pranav Shyam, Girish Sastry, Amanda Askell,
  et~al. 2020{\natexlab{b}}.
\newblock Language models are few-shot learners.
\newblock \emph{Advances in neural information processing systems},
  33:1877--1901.

\bibitem[{Chowdhery et~al.(2022)Chowdhery, Narang, Devlin, Bosma, Mishra,
  Roberts, Barham, Chung, Sutton, Gehrmann et~al.}]{chowdhery2022palm}
Aakanksha Chowdhery, Sharan Narang, Jacob Devlin, Maarten Bosma, Gaurav Mishra,
  Adam Roberts, Paul Barham, Hyung~Won Chung, Charles Sutton, Sebastian
  Gehrmann, et~al. 2022.
\newblock Palm: Scaling language modeling with pathways.
\newblock \emph{arXiv preprint arXiv:2204.02311}.

\bibitem[{Condoravdi et~al.(2003)Condoravdi, Crouch, de~Paiva, Stolle, and
  Bobrow}]{condoravdi-etal-2003-entailment}
Cleo Condoravdi, Dick Crouch, Valeria de~Paiva, Reinhard Stolle, and Daniel~G.
  Bobrow. 2003.
\newblock \href {https://aclanthology.org/W03-0906} {Entailment, intensionality
  and text understanding}.
\newblock In \emph{Proceedings of the {HLT}-{NAACL} 2003 Workshop on Text
  Meaning}, pages 38--45.

\bibitem[{Dagan et~al.(2005)Dagan, Glickman, and Magnini}]{10.1007/11736790_9}
Ido Dagan, Oren Glickman, and Bernardo Magnini. 2005.
\newblock \href {https://doi.org/10.1007/11736790_9} {The pascal recognising
  textual entailment challenge}.
\newblock In \emph{Proceedings of the First International Conference on Machine
  Learning Challenges: Evaluating Predictive Uncertainty Visual Object
  Classification, and Recognizing Textual Entailment}, MLCW'05, page 177–190,
  Berlin, Heidelberg. Springer-Verlag.

\bibitem[{de~Marneffe et~al.(2008)de~Marneffe, Rafferty, and
  Manning}]{de-marneffe-etal-2008-finding}
Marie-Catherine de~Marneffe, Anna~N. Rafferty, and Christopher~D. Manning.
  2008.
\newblock \href {https://aclanthology.org/P08-1118} {Finding contradictions in
  text}.
\newblock In \emph{Proceedings of ACL-08: HLT}, pages 1039--1047, Columbus,
  Ohio. Association for Computational Linguistics.

\bibitem[{Deu{\ss}er et~al.(2023)Deu{\ss}er, Pielka, Pucknat, Jacob,
  Dilmaghani, Nourimand, Kliem, Loitz, Bauckhage, and
  Sifa}]{deusser2023contradiction}
Tobias Deu{\ss}er, Maren Pielka, Lisa Pucknat, Basil Jacob, Tim Dilmaghani,
  Mahdis Nourimand, Bernd Kliem, R{\"u}diger Loitz, Christian Bauckhage, and
  Rafet Sifa. 2023.
\newblock Contradiction detection in financial reports.
\newblock In \emph{Proceedings of the Northern Lights Deep Learning Workshop},
  volume~4.

\bibitem[{Devlin et~al.(2019)Devlin, Chang, Lee, and
  Toutanova}]{devlin-etal-2019-bert}
Jacob Devlin, Ming-Wei Chang, Kenton Lee, and Kristina Toutanova. 2019.
\newblock \href {https://doi.org/10.18653/v1/N19-1423} {{BERT}: Pre-training of
  deep bidirectional transformers for language understanding}.
\newblock In \emph{Proceedings of the 2019 Conference of the North {A}merican
  Chapter of the Association for Computational Linguistics: Human Language
  Technologies, Volume 1 (Long and Short Papers)}, pages 4171--4186,
  Minneapolis, Minnesota. Association for Computational Linguistics.

\bibitem[{Ding et~al.(2023)Ding, Qin, Liu, Chia, Li, Joty, and
  Bing}]{ding-etal-2023-gpt1}
Bosheng Ding, Chengwei Qin, Linlin Liu, Yew~Ken Chia, Boyang Li, Shafiq Joty,
  and Lidong Bing. 2023.
\newblock \href {https://doi.org/10.18653/v1/2023.acl-long.626} {Is {GPT}-3 a
  good data annotator?}
\newblock In \emph{Proceedings of the 61st Annual Meeting of the Association
  for Computational Linguistics (Volume 1: Long Papers)}, pages 11173--11195,
  Toronto, Canada. Association for Computational Linguistics.

\bibitem[{Fortier-Dubois and Rosati(2023)}]{fortier-dubois-rosati-2023-using}
Etienne Fortier-Dubois and Domenic Rosati. 2023.
\newblock \href {https://doi.org/10.18653/v1/2023.acl-short.72} {Using
  contradictions improves question answering systems}.
\newblock In \emph{Proceedings of the 61st Annual Meeting of the Association
  for Computational Linguistics (Volume 2: Short Papers)}, pages 827--840,
  Toronto, Canada. Association for Computational Linguistics.

\bibitem[{Graesser and McMahen(1993)}]{graesser1993anomalous}
Arthur~C Graesser and Cathy~L McMahen. 1993.
\newblock Anomalous information triggers questions when adults solve
  quantitative problems and comprehend stories.
\newblock \emph{Journal of Educational Psychology}, 85(1):136.

\bibitem[{Harabagiu et~al.(2006)Harabagiu, Hickl, and
  Lacatusu}]{10.5555/1597538.1597659}
Sanda Harabagiu, Andrew Hickl, and Finley Lacatusu. 2006.
\newblock Negation, contrast and contradiction in text processing.
\newblock In \emph{Proceedings of the 21st National Conference on Artificial
  Intelligence - Volume 1}, AAAI'06, page 755–762. AAAI Press.

\bibitem[{Hermann et~al.(2015)Hermann, Kocisky, Grefenstette, Espeholt, Kay,
  Suleyman, and Blunsom}]{hermann2015teaching}
Karl~Moritz Hermann, Tomas Kocisky, Edward Grefenstette, Lasse Espeholt, Will
  Kay, Mustafa Suleyman, and Phil Blunsom. 2015.
\newblock Teaching machines to read and comprehend.
\newblock \emph{Advances in neural information processing systems}, 28.

\bibitem[{Hsu et~al.(2021)Hsu, Li, Saez-Trumper, and Hsu}]{9671319}
Cheng Hsu, Cheng-Te Li, Diego Saez-Trumper, and Yi-Zhan Hsu. 2021.
\newblock \href {https://doi.org/10.1109/BigData52589.2021.9671319}
  {Wikicontradiction: Detecting self-contradiction articles on wikipedia}.
\newblock In \emph{2021 IEEE International Conference on Big Data (Big Data)},
  pages 427--436.

\bibitem[{Jin et~al.(2022)Jin, Liu, Liu, and
  Hakkani-Tur}]{jin-etal-2022-improving}
Di~Jin, Sijia Liu, Yang Liu, and Dilek Hakkani-Tur. 2022.
\newblock \href {https://aclanthology.org/2022.sigdial-1.56} {Improving bot
  response contradiction detection via utterance rewriting}.
\newblock In \emph{Proceedings of the 23rd Annual Meeting of the Special
  Interest Group on Discourse and Dialogue}, pages 605--614, Edinburgh, UK.
  Association for Computational Linguistics.

\bibitem[{Ko{\v{c}}isk{\`y} et~al.(2018)Ko{\v{c}}isk{\`y}, Schwarz, Blunsom,
  Dyer, Hermann, Melis, and Grefenstette}]{kovcisky2018narrativeqa}
Tom{\'a}{\v{s}} Ko{\v{c}}isk{\`y}, Jonathan Schwarz, Phil Blunsom, Chris Dyer,
  Karl~Moritz Hermann, G{\'a}bor Melis, and Edward Grefenstette. 2018.
\newblock The narrativeqa reading comprehension challenge.
\newblock \emph{Transactions of the Association for Computational Linguistics},
  6:317--328.

\bibitem[{Lendvai et~al.(2016)Lendvai, Augenstein, Bontcheva, and
  Declerck}]{lendvai-etal-2016-monolingual}
Piroska Lendvai, Isabelle Augenstein, Kalina Bontcheva, and Thierry Declerck.
  2016.
\newblock \href {https://aclanthology.org/L16-1729} {Monolingual social media
  datasets for detecting contradiction and entailment}.
\newblock In \emph{Proceedings of the Tenth International Conference on
  Language Resources and Evaluation ({LREC}'16)}, pages 4602--4605,
  Portoro{\v{z}}, Slovenia. European Language Resources Association (ELRA).

\bibitem[{Lendvai and Reichel(2016)}]{lendvai-reichel-2016-contradiction}
Piroska Lendvai and Uwe Reichel. 2016.
\newblock \href {https://aclanthology.org/W16-5004} {Contradiction detection
  for rumorous claims}.
\newblock In \emph{Proceedings of the Workshop on Extra-Propositional Aspects
  of Meaning in Computational Linguistics ({E}x{P}ro{M})}, pages 31--40, Osaka,
  Japan. The COLING 2016 Organizing Committee.

\bibitem[{Li et~al.(2018)Li, Niu, Al-Doulat, and Park}]{8508308}
Chuqin Li, Xi~Niu, Ahmad Al-Doulat, and Noseong Park. 2018.
\newblock \href {https://doi.org/10.1109/ASONAM.2018.8508308} {A computational
  approach to finding contradictions in user opinionated text}.
\newblock In \emph{2018 IEEE/ACM International Conference on Advances in Social
  Networks Analysis and Mining (ASONAM)}, pages 351--356.

\bibitem[{Mathur et~al.(2022)Mathur, Kunapuli, Bhat, Shrivastava, Manocha, and
  Singh}]{mathur-etal-2022-docinfer}
Puneet Mathur, Gautam Kunapuli, Riyaz Bhat, Manish Shrivastava, Dinesh Manocha,
  and Maneesh Singh. 2022.
\newblock \href {https://doi.org/10.18653/v1/2022.emnlp-main.51} {{D}oc{I}nfer:
  Document-level natural language inference using optimal evidence selection}.
\newblock In \emph{Proceedings of the 2022 Conference on Empirical Methods in
  Natural Language Processing}, pages 809--824, Abu Dhabi, United Arab
  Emirates. Association for Computational Linguistics.

\bibitem[{Merity et~al.(2016)Merity, Xiong, Bradbury, and
  Socher}]{merity2016pointer}
Stephen Merity, Caiming Xiong, James Bradbury, and Richard Socher. 2016.
\newblock \href {http://arxiv.org/abs/1609.07843} {Pointer sentinel mixture
  models}.

\bibitem[{Mündler et~al.(2023)Mündler, He, Jenko, and
  Vechev}]{mündler2023selfcontradictory}
Niels Mündler, Jingxuan He, Slobodan Jenko, and Martin Vechev. 2023.
\newblock \href {http://arxiv.org/abs/2305.15852} {Self-contradictory
  hallucinations of large language models: Evaluation, detection and
  mitigation}.

\bibitem[{Nie et~al.(2021)Nie, Williamson, Bansal, Kiela, and
  Weston}]{nie-etal-2021-like}
Yixin Nie, Mary Williamson, Mohit Bansal, Douwe Kiela, and Jason Weston. 2021.
\newblock \href {https://doi.org/10.18653/v1/2021.acl-long.134} {{I} like fish,
  especially dolphins: Addressing contradictions in dialogue modeling}.
\newblock In \emph{Proceedings of the 59th Annual Meeting of the Association
  for Computational Linguistics and the 11th International Joint Conference on
  Natural Language Processing (Volume 1: Long Papers)}, pages 1699--1713,
  Online. Association for Computational Linguistics.

\bibitem[{OpenAI(2022)}]{OpenAI_ChatGPT_Blog}
OpenAI. 2022.
\newblock \href {https://openai.com/blog/chatgpt/} {Chatgpt: Optimizing
  language models for dialogue}.
\newblock OpenAI Blog.

\bibitem[{OpenAI(2023)}]{gpt4openai}
OpenAI. 2023.
\newblock Gpt-4 technical report.
\newblock \emph{arXiv preprint arXiv:2303.08774}.

\bibitem[{Otero and Kintsch(1992)}]{doi:10.1111/j.1467-9280.1992.tb00034.x}
José Otero and Walter Kintsch. 1992.
\newblock \href {https://doi.org/10.1111/j.1467-9280.1992.tb00034.x} {Failures
  to detect contradictions in a text: What readers believe versus what they
  read}.
\newblock \emph{Psychological Science}, 3(4):229--236.

\bibitem[{Pangakis et~al.(2023)Pangakis, Wolken, and
  Fasching}]{Pangakis2023AutomatedAW}
Nicholas Pangakis, Samuel Wolken, and Neil Fasching. 2023.
\newblock \href {https://api.semanticscholar.org/CorpusID:259000016} {Automated
  annotation with generative ai requires validation}.
\newblock \emph{ArXiv}, abs/2306.00176.

\bibitem[{Rosemblat et~al.(2019)Rosemblat, Fiszman, Shin, and
  Kilicoglu}]{ROSEMBLAT2019103275}
Graciela Rosemblat, Marcelo Fiszman, Dongwook Shin, and Halil Kilicoglu. 2019.
\newblock \href {https://doi.org/https://doi.org/10.1016/j.jbi.2019.103275}
  {Towards a characterization of apparent contradictions in the biomedical
  literature using context analysis}.
\newblock \emph{Journal of Biomedical Informatics}, 98:103275.

\bibitem[{Sarafraz(2012)}]{sarafraz2012finding}
Farzaneh Sarafraz. 2012.
\newblock \emph{Finding conflicting statements in the biomedical literature}.
\newblock Ph.D. thesis, The University of Manchester (United Kingdom).

\bibitem[{Schuster et~al.(2022{\natexlab{a}})Schuster, Chen, Buthpitiya,
  Fabrikant, and Metzler}]{schuster2022stretching}
Tal Schuster, Sihao Chen, Senaka Buthpitiya, Alex Fabrikant, and Donald
  Metzler. 2022{\natexlab{a}}.
\newblock \href {http://arxiv.org/abs/2204.07447} {Stretching sentence-pair nli
  models to reason over long documents and clusters}.

\bibitem[{Schuster et~al.(2022{\natexlab{b}})Schuster, Chen, Buthpitiya,
  Fabrikant, and Metzler}]{schuster-etal-2022-stretching}
Tal Schuster, Sihao Chen, Senaka Buthpitiya, Alex Fabrikant, and Donald
  Metzler. 2022{\natexlab{b}}.
\newblock \href {https://doi.org/10.18653/v1/2022.findings-emnlp.28}
  {Stretching sentence-pair {NLI} models to reason over long documents and
  clusters}.
\newblock In \emph{Findings of the Association for Computational Linguistics:
  EMNLP 2022}, pages 394--412, Abu Dhabi, United Arab Emirates. Association for
  Computational Linguistics.

\bibitem[{Singhal et~al.(2023)Singhal, Tu, Gottweis, Sayres, Wulczyn, Hou,
  Clark, Pfohl, Cole-Lewis, Neal, Schaekermann, Wang, Amin, Lachgar, Mansfield,
  Prakash, Green, Dominowska, y~Arcas, Tomasev, Liu, Wong, Semturs, Mahdavi,
  Barral, Webster, Corrado, Matias, Azizi, Karthikesalingam, and
  Natarajan}]{singhal2023expertlevel}
Karan Singhal, Tao Tu, Juraj Gottweis, Rory Sayres, Ellery Wulczyn, Le~Hou,
  Kevin Clark, Stephen Pfohl, Heather Cole-Lewis, Darlene Neal, Mike
  Schaekermann, Amy Wang, Mohamed Amin, Sami Lachgar, Philip Mansfield, Sushant
  Prakash, Bradley Green, Ewa Dominowska, Blaise~Aguera y~Arcas, Nenad Tomasev,
  Yun Liu, Renee Wong, Christopher Semturs, S.~Sara Mahdavi, Joelle Barral,
  Dale Webster, Greg~S. Corrado, Yossi Matias, Shekoofeh Azizi, Alan
  Karthikesalingam, and Vivek Natarajan. 2023.
\newblock \href {http://arxiv.org/abs/2305.09617} {Towards expert-level medical
  question answering with large language models}.

\bibitem[{Sun et~al.(2022)Sun, He, Qiu, and Huang}]{sun-etal-2022-bertscore}
Tianxiang Sun, Junliang He, Xipeng Qiu, and Xuanjing Huang. 2022.
\newblock \href {https://doi.org/10.18653/v1/2022.emnlp-main.245} {{BERTS}core
  is unfair: On social bias in language model-based metrics for text
  generation}.
\newblock In \emph{Proceedings of the 2022 Conference on Empirical Methods in
  Natural Language Processing}, pages 3726--3739, Abu Dhabi, United Arab
  Emirates. Association for Computational Linguistics.

\bibitem[{Sun et~al.(2023)Sun, Li, Li, Wu, Guo, Zhang, and Wang}]{sun2023text}
Xiaofei Sun, Xiaoya Li, Jiwei Li, Fei Wu, Shangwei Guo, Tianwei Zhang, and
  Guoyin Wang. 2023.
\newblock \href {http://arxiv.org/abs/2305.08377} {Text classification via
  large language models}.

\bibitem[{Touvron et~al.(2023)Touvron, Martin, Stone, Albert, Almahairi,
  Babaei, Bashlykov, Batra, Bhargava, Bhosale et~al.}]{touvron2023llama}
Hugo Touvron, Louis Martin, Kevin Stone, Peter Albert, Amjad Almahairi, Yasmine
  Babaei, Nikolay Bashlykov, Soumya Batra, Prajjwal Bhargava, Shruti Bhosale,
  et~al. 2023.
\newblock Llama 2: Open foundation and fine-tuned chat models.
\newblock \emph{arXiv preprint arXiv:2307.09288}.

\bibitem[{Wang et~al.(2023)Wang, Lyu, Ji, Zhang, Yu, Shi, and
  Tu}]{wang2023documentlevel}
Longyue Wang, Chenyang Lyu, Tianbo Ji, Zhirui Zhang, Dian Yu, Shuming Shi, and
  Zhaopeng Tu. 2023.
\newblock \href {http://arxiv.org/abs/2304.02210} {Document-level machine
  translation with large language models}.

\bibitem[{Wang et~al.(2021)Wang, Liu, Xu, Zhu, and Zeng}]{Wang2021WantTR}
Shuohang Wang, Yang Liu, Yichong Xu, Chenguang Zhu, and Michael Zeng. 2021.
\newblock \href {https://api.semanticscholar.org/CorpusID:237363383} {Want to
  reduce labeling cost? gpt-3 can help}.
\newblock \emph{ArXiv}, abs/2108.13487.

\bibitem[{Wu et~al.(2022)Wu, Niu, and Rahman}]{10.1145/3477495.3531881}
Xiangcheng Wu, Xi~Niu, and Ruhani Rahman. 2022.
\newblock \href {https://doi.org/10.1145/3477495.3531881} {Topological analysis
  of contradictions in text}.
\newblock In \emph{Proceedings of the 45th International ACM SIGIR Conference
  on Research and Development in Information Retrieval}, SIGIR '22, page
  2478–2483, New York, NY, USA. Association for Computing Machinery.

\bibitem[{Yin et~al.(2021{\natexlab{a}})Yin, Radev, and Xiong}]{yin2021docnli}
Wenpeng Yin, Dragomir Radev, and Caiming Xiong. 2021{\natexlab{a}}.
\newblock \href {http://arxiv.org/abs/2106.09449} {Docnli: A large-scale
  dataset for document-level natural language inference}.

\bibitem[{Yin et~al.(2021{\natexlab{b}})Yin, Radev, and
  Xiong}]{yin-etal-2021-docnli}
Wenpeng Yin, Dragomir Radev, and Caiming Xiong. 2021{\natexlab{b}}.
\newblock \href {https://doi.org/10.18653/v1/2021.findings-acl.435}
  {{D}oc{NLI}: A large-scale dataset for document-level natural language
  inference}.
\newblock In \emph{Findings of the Association for Computational Linguistics:
  ACL-IJCNLP 2021}, pages 4913--4922, Online. Association for Computational
  Linguistics.

\bibitem[{Zhang et~al.(2023)Zhang, Ladhak, Durmus, Liang, McKeown, and
  Hashimoto}]{zhang2023benchmarking}
Tianyi Zhang, Faisal Ladhak, Esin Durmus, Percy Liang, Kathleen McKeown, and
  Tatsunori~B. Hashimoto. 2023.
\newblock \href {http://arxiv.org/abs/2301.13848} {Benchmarking large language
  models for news summarization}.

\bibitem[{Zheng et~al.(2022)Zheng, Zhou, Zheng, Peng, Guo, Wu, Niu, Wu, and
  Huang}]{zheng-etal-2022-cdconv}
Chujie Zheng, Jinfeng Zhou, Yinhe Zheng, Libiao Peng, Zhen Guo, Wenquan Wu,
  Zheng-Yu Niu, Hua Wu, and Minlie Huang. 2022.
\newblock \href {https://doi.org/10.18653/v1/2022.emnlp-main.2} {{CDC}onv: A
  benchmark for contradiction detection in {C}hinese conversations}.
\newblock In \emph{Proceedings of the 2022 Conference on Empirical Methods in
  Natural Language Processing}, pages 18--29, Abu Dhabi, United Arab Emirates.
  Association for Computational Linguistics.

\end{thebibliography}
\bibliographystyle{acl_natbib}

\newpage
\appendix

\section{Dataset Details}
\label{sec:dataset_details}

We use three publically available datasets covering different domains to build \methodnosp. More specifically, we use the following datasets:

\begin{itemize}
\setlength\itemsep{-0.2em}
    \item \textbf{News Articles}: CNN-DailyMail dataset \cite{hermann2015teaching}, an open-source corpus of 93k articles from CNN and 220k articles from Daily Mail and collect 158 documents for \dataposnosp.
    \item \textbf{Stories}: NarrativeQA \cite{kovcisky2018narrativeqa}, which is an open-source question-answering dataset and consists of 1,572 stories and their human-generated summaries. We collected 141 summaries for \dataposnosp.
    \item \textbf{Wikipedia}: WikiText \cite{merity2016pointer}, an open-source language modelling dataset containing verified Wikipedia documents and select 150 documents for \datapos.
\end{itemize} 
We release our dataset under Apache 2.0 license \footnote{\url{https://www.apache.org/licenses/LICENSE-2.0}}.

\section{Model details}
\label{sec:model_details}

We use the following state-of-the-art LLMs to test both open-source and closed-source models in a zero-shot setting on \method.
\begin{itemize}[itemsep=-1pt]
    \item \textbf{GPT3.5}: Also called ChatGPT\footnote{\url{https://openai.com/blog/chatgpt}}, this is an improved version of GPT3 \cite{brown2020language} optimized for chat. We use the \textit{gpt-3.5-turbo-0613} model from the OpenAI API\footnote{\url{https://api.openai.com/}}.
    \item \textbf{GPT4} \cite{gpt4openai}: GPT4 is the latest iteration of the GPT models and is also optimized for chat. We use the \textit{gpt-4-0613} model from the OpenAI API.
    \item \textbf{PaLM2} \cite{anil2023palm}: We use the PaLM 2 model (\textit{text-bison}) from the Vertex AI platform from Google Cloud\footnote{\url{https://cloud.google.com/vertex-ai/docs/generative-ai/learn/models}}. 
    \item \textbf{LLaMAv2} \cite{touvron2023llama}: We use the \textit{Llama-2-Chat-70B} model for our experiments. We used the best performing model that is fine-tuned on dialog data to follow 0-shot instruction. 
\end{itemize}

Unless otherwise specified, we use the default configurations and decoding parameters for all our experiments.



\section{Questions for Annotation}
\label{sec:anno}
We highlight the original statement as well as the introduced
self-contradiction in the document as 1 for annotators2
to verify the validity of document-level self-contradiction. Annotators, guided by comprehensive guidelines, were tasked with the following questions:
\begin{enumerate}[label=Q\arabic*.,itemsep=-1pt]
    \item Do you think the two statements contradict each other?
    \item (If applicable): Is the position of the inserted statement (red color) feasible?
    \item Overall, do you think it makes an acceptable contradictory document? 
    \item How close in the context of the modified sentence can you find the evidence for the self-contradiction? (As described in \ref{sec:scope})
    \item Select Type(s) of self-contradiction.
\end{enumerate}
Each modified document was evaluated by two annotators, establishing validity through consensus on the self-contradiction and document validity. For Q2, if an alternative insertion place is given by the annotators, we add this modification as another contradictory document in our setting. 

Examples are filtered if both annotators answered ``Yes'' for Q1, Q2, and Q3. For Q4, 88\% of the annotators agree with each other, and for 12\% that do not agree, we select the ``closer one'' as the final tag. For Q5, we combine all types selected by both annotators.

To verify the annotation quality, we run another expert filter by the authors of this work to verify controversial cases marked by annotators. Regarding the self-contradiction injection method, the final \method contains 271 documents created by contradictory statement replacing and 178 documents created by contradictory statement inserting.
\section{Prompts for experiment setting}
\label{sec:prompts}

For evaluating the different LLMs on \method, we set up three experiments. Here, we provide the corresponding prompts for each of the experimental settings.

\begin{figure*}
    \centering
    \includegraphics[width=1.05\textwidth]{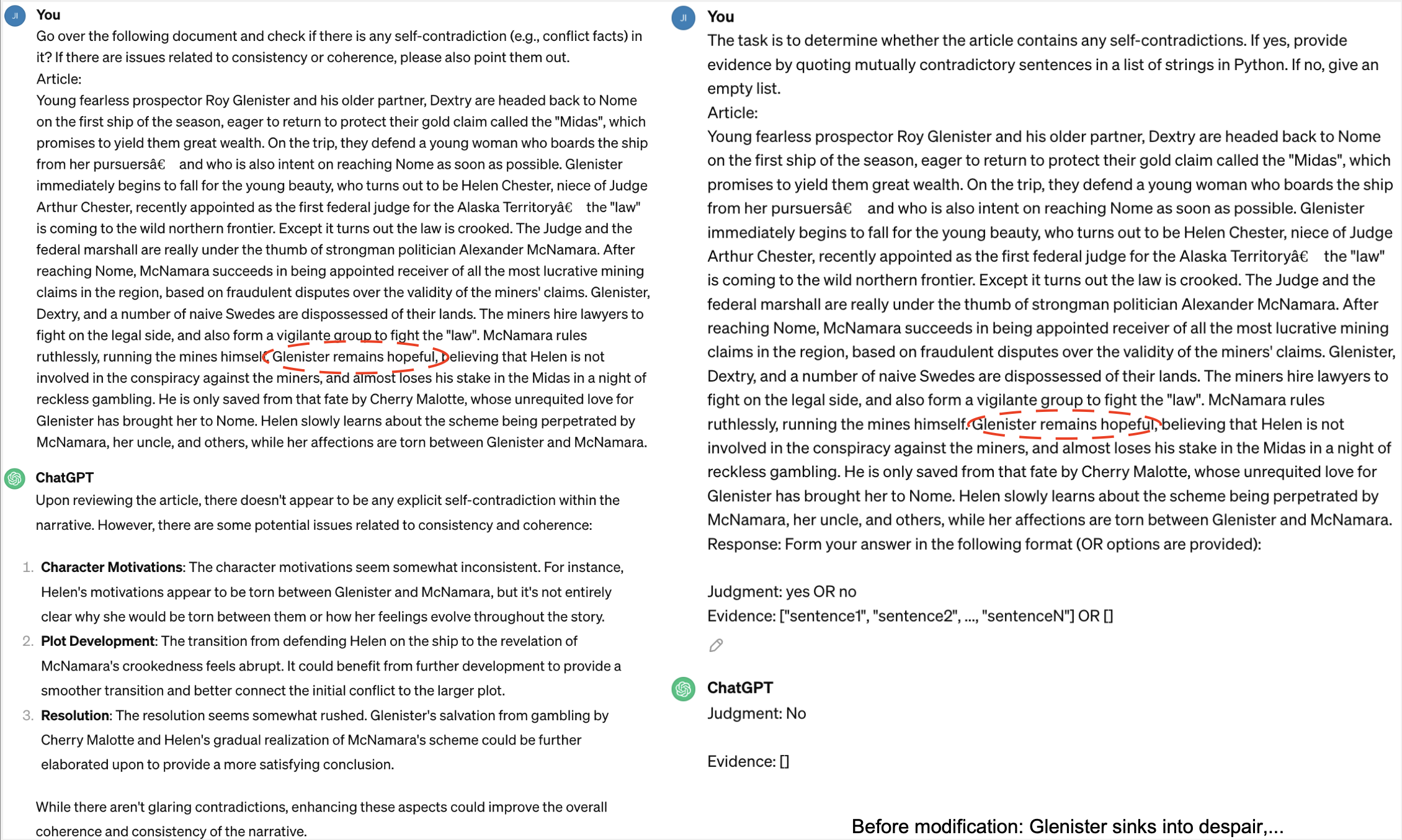}
    \caption{Comparison between free-format evaluation and judge-then-find evaluation on GPT-3.5. The emotion of the character contradicts the context, and is thus marked as ``Emotion/Mood/Feeling'' self-contradiction.}
    \label{fig:eval_case}
\end{figure*}

\begin{itemize}
    \item \textbf{Binary Judgment Prompt} 
    
    \textbf{[Insert Document here]} 
    
    \textit{Determine whether the given document contains any self-contradictions. Only answer "yes" or "no"!}
    \item \textbf{Self-Contradiction in Top $k$ Prompt:} 
    
    \textit{Self-Contradictory Article: An article is deemed self-contradictory when it contains one(self-conflict mention) or more statements that conflict with each other, making them mutually exclusive. The following article contains one self-contradiction. The task is to find where it is. Provide evidence by quoting mutually contradictory sentences from the article. Article: }
    
    \textbf{[Insert Document here]} 
    
    \textit{Please respond by giving the five most likely sentences that can reflect article-level contradiction(s), ranked by high to low possibility. Don't explain.}
    
    \item \textbf{Judgment then Find Prompt:} 
    
    \textit{The task is to determine whether the article contains any self-contradictions. If yes, provide evidence by quoting mutually contradictory sentences in a list of strings in Python. If no, give an empty list.} 
    
    \textbf{[Insert Document here]} 
    
    \textit{Response: Form your answer in the following format (OR options are provided):} 
    
    \textit{Judgment: yes OR no}
    
    \textit{Evidence: ["sentence1", "sentence2", ..., "sentenceN"] OR []}

    \item \textbf{Prompt for Effect of Prompts experiment:}

    \textit{Go over the following document and check if there is any self-contradiction (e.g., conflict facts) in it? If there are issues related to consistency or coherence, please also point them out.} 
    
    Figure \ref{fig:eval_case} compares the GPT-3.5 outputs on this prompt (free-format evaluation) and the judge-then-find evaluation.
\end{itemize}

\end{document}